\documentclass[12pt,preprint]{elsarticle}

\usepackage{amsfonts}
\usepackage{amssymb}
\usepackage{amsmath}
\usepackage{amsthm}
\usepackage{enumerate}
\usepackage{color}
\usepackage{hyperref}
\usepackage{tikz}

 \newtheorem{thm}{Theorem}[section]

 \newtheorem{prop}[thm]{Proposition}
 \theoremstyle{definition}
 \newtheorem{defn}[thm]{Definition}

 \theoremstyle{remark}
 \newtheorem{rem}[thm]{Remark}
 
 \newtheorem{exam}[thm]{Example}

\newcommand{\B}{\boldsymbol{\cdot}}

\begin{document}
\begin{frontmatter}

\title{A fusion method for multi-valued data}
\author[Martin]{Martin Pap\v{c}o}
\author[UPNA]{Iosu Rodr\'\i guez-Mart\'inez}
\author[UPNA]{Javier Fumanal-Idocin}
\author[KAU]{Abdulrahman H. Altalhi}
\author[UPNA,KAU]{Humberto Bustince}
\address[Martin]{Catholic University in Ru\v{z}omberok, Hrabovsk\'{a} cesta~1,
034~01~Ru\v{z}omberok, Slovak Republic and Mathematical Institute,
Slovak Academy of Sciences, \v{S}tef\'{a}nikova 49, 814~73~Bratislava\\
e-mail: papco@ruzomberok.sk}

\address[UPNA]{Department of Statistics, Computer Science and Mathematics, Universidad P\'ublica de Navarra, Campus Arrosadia s/n, 31006 Pamplona\\
email: iosu.rodriguez@unavarra.es}

\address[KAU]{Faculty of Computing and Information Technology, King Abdulaziz University, 80203, Jeddah 21589, Saudi Arabia\\
email: ahaltalhi@kau.edu.sa}

%
%
%

\begin{abstract}
In this paper we propose an extension of the notion of deviation-based aggregation function tailored to aggregate multidimensional data. Our objective is both to improve the results obtained by other methods that try to select the best aggregation function for a particular set of data, such as penalty functions, and to reduce the temporal complexity required by such approaches. We discuss how this notion can be defined and present three illustrative examples of the applicability of our new proposal in areas where temporal constraints can be strict, such as image processing, deep learning and decision making, obtaining favourable results in the process.
\end{abstract}


\end{frontmatter}

\section{Introduction}
In many different fields of human activity it is necessary to fuse information. More specifically, a common task is to consider a set of data (called inputs) and to replace them by one single value, the output, which is assumed to represent all of them. In order to develop this procedure, one widely used tool is that of aggregation functions, understood as increasing functions which satisfy suitable boundary conditions, described more in depth in~\cite{BeliakovBustince, BeliakovPradera}.

Aggregation functions have attracted the interest of a large number of researchers, both from theoretical and applied points of view~\cite{BeliakovBustince, BeliakovPradera, LuccaCF, GracalizFernandez}. In particular, large efforts have been devoted to determine which is the best aggregation function to be used in a given setting~\cite{CalvoBeliakov,YagerRybalov}. In this sense, Yager et al.~\cite{Yager} introduced the notion of penalty function as a way to determine how dissimilar a possible output is from a set of inputs, and showed how, by means of a minimization procedure, this penalty function can be used to define an appropriate aggregation function. That is, the penalty function can be used to determine which is the output that best represents a given set of inputs. The suggested approach was elaborated in sequel papers, e.g. in~\cite{CalvoBeliakov,CalvoMesiar}.

Even if these functions have been successfully applied in image processing~\cite{Paternain} or decision making~\cite{consensus,ElkanoGalar,GalarFernandez} the fulfillment of the minimization procedure requires of quite strict convexity and differentiability assumption, which leads to technical difficulties in order to apply this notion~\cite{BustinceBeliakov}. For this reason, the attention of some researchers has been recently directed to the notion of deviation-based aggregation functions~\cite{Decky}. This idea is based on a search for some sort of an equilibrium (the equation solution) described in terms of Dar\'oczy means ~\cite{Daroczy}. However, in general, such means are not monotone, so in order to recover aggregation functions, some additional study was needed and done in~\cite[Definition~3.4, Theorem~3.1, Theorem~4.1, Definition~4.1]{Decky} via ``inf-sup-middle-approach''. Namely, the representative output is chosen so that it lies in the middle of $$\inf\{y\,|\,\hspace{-60pt}\sum\limits_{\hspace{60pt}\text{throughout all inputs}}\hspace{-60pt} D(\text{input},y)<0\}\quad\text{and}\quad \sup\{y\,|\,\hspace{-60pt}\sum\limits_{\hspace{60pt}\text{throughout all inputs}}\hspace{-60pt} D(\text{input},y)>0\},$$ where $D$ is the function used to measure the deviation of the output from the inputs.

So far, all the discussed developments were done for numerical (unidimensional) inputs. But in many cases, inputs to be fused are or can be considered as multidimensional, and the expected output should also be multidimensional. This is the case, for instance, in color images, where the intensity of each pixel can be given, e.g, in terms of a tuple $(R,G,B)$, where $R$ denotes the intensity of the pixel in the red channel, $G$, in the green channel, and $B$, in the blue channel. Similarly, in deep learning, the data we deal with are given either in the format of color images or three dimensional feature images, that can be thought of as generalizations of the first ones with $N$ channels instead of just 3. It is also the case in multi-expert multicriteria decision making problems~\cite{Herrera}, where, if we have $n$ experts and if we fix two alternatives $i$ and  $j$, then the set of preferences given by all the experts for the alternative $i$ over the alternative $j$ can be seen as an $n$-dimensional vector.

All these problems work with matrices of multidimensional data, and therefore our goal in this paper is to extend the notion of deviation functions to this setting. More specifically, we are going to build a method based on appropriate multidimensional deviation functions, in order to replace related subsets of these data by a single representative. We take into account the fact that some of these problems, such as deep learning, work with real valued data instead of values restricted to the unit interval, so our method is prepared to work with values in the real domain.

In order to show the usefulness of this approach, we are going to consider three specific illustrative examples in the aforementioned settings: image processing, where the application of techniques such as penalty functions is restricted to scenarios that work with small image sizes and few aggregation functions; deep learning, where approaches based on the use of penalty  functions are intractable due to the huge computational complexity; and decision making, where the a priori choice of several appropriate aggregation functions can be difficult.

The paper is structured in the following manner: Crucial notions are recalled in Section~\ref{Preliminaries_section}. Section~\ref{MDBA_section}, the core of the paper, is devoted to the construction of a weighted deviation-based matrix aggregation. Then the approach is illustrated in Section~\ref{Exam_section} via some examples which, at the same time, indicate possible applications. Finally, some remarks conclude the contribution. 

\section{Preliminaries}\label{Preliminaries_section}

In this section, we review some concepts which are relevant for the subsequent developments of our work. We start recalling the notion of aggregation function.

In what follows the chain of natural numbers $\{1,2,\dots,m\}$ will be denoted by $[m]$.

\begin{defn}
Let $I$ be a closed interval of real numbers, and let $a=\min I$ and $b=\max I$. An ($n$-dimensional) aggregation function on the interval $I$ is an increasing mapping $M:I^n \to I$ such that $M(a,\dots,a)=a$, $M(b,\dots,b)=b$.
\end{defn}

An aggregation function $M:I^n \to I$ such that, for every $x_1,\dots,x_n \in I$ it holds that:
\[
\min (x_1,\dots,x_n) \le M(x_1,\dots,x_n) \le \max (x_1\dots,x_n)
\]
is called averaging~\cite{BeliakovBustince}. Note that an averaging aggregation function is idempotent, i.e., $M(x,\dots,x)=x$ for every $x \in I$. In fact, due to monotonicity, an idempotent aggregation function is also averaging.

Among the most widely known examples of aggregation functions we can mention, for instance, the arithmetic mean, the minimum, the product or the maximum. See~\cite{BeliakovPradera} for a comprehensive review on the subject.

The main goal of this work, as we have said, is to extend the notion of deviation function to a multidimensional setting. We start recalling the definition of this concept in the real case.

\begin{defn}\label{defn_dev_function}
Let $I$ be a closed interval of real numbers. Then a mapping $D\!:\ I^2\rightarrow\mathbb{R}$ is said to be a \emph{deviation function} if and only if
\begin{enumerate}
  \item for every $x\in I$, $D(x,\cdot)\!:\ I\rightarrow\mathbb{R}$ is continuous and strictly increasing;
  \item $D(x,x)=0$ for every $x\in I$.
\end{enumerate}
\end{defn}

\begin{exam}
For any closed interval of real numbers $I$, the function $D(x,y)=y-x$ is a deviation function.
\end{exam}

From the point of view of this paper, the interest of deviation functions lays on the fact that they can be used to define a specific class of means, known as Dar\'oczy means.

\begin{defn}\label{defn_Daroczy_mean}
Let $I$ be a closed interval of real numbers, let $D\!:\ I^2\rightarrow\mathbb{R}$ be a given deviation function, and let $n$ be a natural number. Then a mapping $M_D\!:\ I^n\rightarrow\mathbb{R}$ is said to be a \emph{Dar\'oczy mean} if and only if $M_D$ is given by
$$M_D(\mathbf{x})=y,$$
where $y$ is the solution of the equation
$$\sum\limits_{i=1}^n D(x_i,y)=0.$$
\end{defn}

Note that in Definition~\ref{defn_Daroczy_mean}, for each possible real value $y$, the deviation functions are used as a tool to determine how different each input $x_i$ is from $y$, and the final output of the function $M_D$ is precisely that $y$ such that the sum of all these differences is as small as possible. However, in order to get a better value to fuse all the inputs, it is advisable to consider a stricter definition of deviation function, leading to the notion of moderate deviation function that we discuss now.

\begin{defn}[Improvement of Definition~\ref{defn_dev_function}; see~\cite{Decky}]\label{defn_mod_dev_function}
Let $I$ be a closed interval of real numbers. Then a mapping $D\!:\ I^2\rightarrow\mathbb{R}$ is said to be a \emph{moderate deviation function} if and only if
\begin{enumerate}
  \item for every $x\in I$, $D(x,\cdot)\!:\ I\rightarrow\mathbb{R}$ is non-decreasing;
  \item for every $y\in I$, $D(\cdot,y)\!:\ I\rightarrow\mathbb{R}$ is non-increasing;
  \item $D(x,y)=0$ if and only if $x=y$, $x,\,y\in I$.
\end{enumerate}
\end{defn}

Observe that with this new definition, if all the inputs are equal to $x$ and if we take as definition of $M_D$ the same as in Definition~\ref{defn_Daroczy_mean}, we have that the resulting output is equal to $x$, too.

The prototypical example of moderate deviation function is provided by the function $D(x,y)=y-x$. Based on this function, we can introduce a general family of moderate deviation functions, as follows.

\begin{prop}\label{prop_basic_MDF}
Let $f\!:\ \mathbb{R}\rightarrow\mathbb{R}$ be a non-decreasing function such that $f(x)=0$ if and only if $x=0$. Let $s\!:\ [0,1]\rightarrow\mathbb{R}$ be a strictly increasing function. Then a mapping $D_{f,s}\!:\ [0,1]^2\rightarrow\mathbb{R}$ given by
$$D_{f,s}(x,y)=f\big(s(y)-s(x)\big)$$
is a moderate deviation function. It is called a \emph{basic moderate deviation function}.
\end{prop}

\noindent{\sl Proof.} It follows from a straightforward calculation.

It is worth mentioning that via deviation-based approaches there is possible to catch several types of mixture operators and their generalizations (more in~\cite{RibeiroPereira}, \cite{Spirkova}, \cite{SpirkovaBeliakov}, \cite{SpirkovaKral}, \cite{Stupnanova}).

\section{New fusion method: Multidimensional deviation-based aggregation}\label{MDBA_section}

Let~$\mathbb{A}$ be a $p\times q$ matrix of real $n$-tuples (vectors) such that both $p$ and $q$ are divisible by a positive integer $r$, $r\neq 1$. Clearly, the matrix~$\mathbb{A}$ can be considered as composed of $n$ submatrices $\mathbb{A}_1,\, \mathbb{A}_2,\, \dots,\, \mathbb{A}_n$ with real entries, where $\mathbb{A}_j$ is the matrix that we obtain by considering just the $j$-th component of the elements in~$\mathbb{A}$. In this way,  we can identify the matrix~$\mathbb{A}$ with the vector $(\mathbb{A}_1,\mathbb{A}_2,\dots,\mathbb{A}_n)$ of matrices.

Now, if we decompose the matrix~$\mathbb{A}$ into $(p/r)\cdot(q/r)$ ``mutually disjoint'' $r\times r$ submatrices $\mathbb{B}^{\alpha\beta}$, $\alpha\in\{1,2,\dots,p/r\}$, $\beta\in\{1,2,\dots,q/r\}$, we can replace each such submatrix of $n$-tuples by one appropriate representative which is again a $n$-tuple, leading to a $(p/r)\times (q/r)$ matrix $\mathbb{C}$ of real $n$-tuples (referring to the previous remark: aggregated ``RGB''-estimations of the three experts). As we have already discussed, this abstract procedure corresponds actually to different information fusion processes that are carried out in applications such as image processing, deep learning or decision making.


We are going now to formalize this procedure. First of all, let us denote $\mathcal{I}=[p/r]$ and $\mathcal{J}=[q/r]$. If we denote the element in position $(r,s)$ of~$\mathbb{A}$ by $a_{rs}$, then for every $\alpha\in\mathcal{I}$ and $\beta\in\mathcal{J}$, we can define a $r\times r$ matrix $\mathbb{B}^{\alpha\beta}$ of $n$-tuples as
\begin{equation}
b_{i,j}^{\alpha\beta}=a_{(\alpha-1)r+i,(\beta-1)r+j}\quad \text{for}\quad i,\, j\in [r].\label{b_alphabeta}
\end{equation}
In this way, the matrix~$\mathbb{A}$ can be seen as a ``disjoint union'' of matrices $\mathbb{B}^{\alpha\beta}$ over all indices $\alpha\in\mathcal{I}$, $\beta\in\mathcal{J}$.

Let $\alpha$ be an arbitrary but fixed index from $\mathcal{I}$ and $\beta$ from $\mathcal{J}$, respectively. Then the matrix $\mathbb{B}^{\alpha\beta}$ consists from $n$ of $r\times r$ square matrices
\begin{equation}
\mathbb{B}_k^{\alpha\beta} =
\left(
\begin{array}{cccc}
b_{11k}^{\alpha\beta}&b_{12k}^{\alpha\beta}&\cdots &b_{1rk}^{\alpha\beta}\\[6pt]
b_{21k}^{\alpha\beta}&b_{22k}^{\alpha\beta}&\cdots &b_{2rk}^{\alpha\beta}\\[6pt]
\vdots&\vdots&\ddots&\vdots\\[6pt]
b_{r1k}^{\alpha\beta}&b_{r2k}^{\alpha\beta}&\cdots &b_{rrk}^{\alpha\beta}\\
\end{array}
\right)\ \text{for every}\ k\in[n].
\label{B_alphabeta}
\end{equation}
Note that, from the geometrical point of view, if we consider that the elements in~$\mathbb{A}$ (and hence in $\mathbb{B}^{\alpha\beta}$) are points in the $n$-dimensional Euclidean space $\mathbf{E}^n$, then elements of the matrix $\mathbb{B}_k^{\alpha\beta}$ correspond to the projections onto the $k$-th axis.

Our next step is to aggregate all these values for every $k\in[n]$ using a suitable deviation-based aggregation function. In order to choose an appropriate deviation function, and specially taking into account possible applications, it is necessary to fix a closed real interval to be used as a domain of a deviation-based aggregation function in question. For this reason, let
$$\wedge_k^{\alpha\beta}=\min\limits_{i,j\in[r]}\{b_{ijk}^{\alpha\beta}\}\quad
\text{and}\quad
\vee_k^{\alpha\beta}=\max\limits_{i,j\in[r]}\{b_{ijk}^{\alpha\beta}\},\quad k\in{n}.$$
Denote by $I_k^{\alpha\beta}$ the closed real interval $[\wedge_k^{\alpha\beta},\vee_k^{\alpha\beta}]$, i.e. \begin{equation}
I_k^{\alpha\beta}=\{x\in\mathbb{R}\,|\,\wedge_k^{\alpha\beta}\leq x\leq\vee_k^{\alpha\beta}\}.
\label{I_k}
\end{equation}
It will serve as the domain of a deviation-based aggregation function which will be applied as the aggregation of $\mathbb{B}_k^{\alpha\beta}$ elements.

Now we describe the final step in order to get a $(p/r)\times (q/r)$ matrix $\mathbb{C}$ to replace the original $p\times q$ matrix~$\mathbb{A}$, in such a way that the elements of~$\mathbb{C}$ are $n$-tuples obtained by fusing the elements of an appropriate submatrix of~$\mathbb{A}$.

\begin{thm}
Let $I$ be an interval of real numbers. Let~$\mathbb{X}$ be a $s\times t$ matrix such that $x_{ij}\in I$ for every $i\in[s]$, $j\in[t]$. Denote by $I_\mathbb{X}$ a closed real interval $$\big[\min_{i\in[s],j\in[t]}\{x_{ij}\}, \max_{i\in[s],j\in[t]}\{x_{ij}\}\big].$$
Let $D\!:\ I_\mathbb{X}\times I_\mathbb{X}\rightarrow\mathbb{R}$ be a moderate deviation function. Define a function $M_D\!:\ I_\mathbb{X}^{s\cdot t}\rightarrow \mathbb{R}$ by
\begin{align}
M_D(\mathbb{X})=
\frac{1}{2}
\big(\sup&\big\{y\in I\mid\sum_{i=1}^{s}\sum_{j=1}^{t}D(x_{ij},y)<0\big\}+\nonumber\\
+\inf&\big\{y\in I\mid\sum_{i=1}^{s}\sum_{j=1}^{t}D(x_{ij},y)>0\big\}\big).\tag{MDX}\label{MDX}
\end{align}
Then $M_D$ is an idempotent symmetric aggregation function.
\end{thm}

Note that the previous expression is an extension of the notion of D-mean presented in~\cite{Decky} for the case of inputs~$\mathbb{X} \in I^{p \cdot q}$.

\begin{proof}
The theorem can be proved by rewriting of the Theorem~3.1 proof in~\cite{Decky} as follows. Since, for every $i\in[s]$, $j\in[t]$, the element $x_{ij}$ of $\mathbb{X}$ belongs to the closed interval $I_\mathbb{X}$ one can consider the matrix $\mathbb{X}$ as an $(s\cdot t)$-tuple of real numbers, i.e. a real vector $\mathbf{x}$ for which the construction in question is applied. At the same time it is necessary to replace unit interval $[0,1]$ as the domain of the corresponding moderate deviation function by $I_\mathbb{X}$, or its boundary points by
$\min_{i\in[s],j\in[t]}\{x_{ij}\}$ and $\max_{i\in[s],j\in[t]}\{x_{ij}\}$, respectively.
\end{proof}

The fusion procedure can be done by assigning a different weight to the different considered inputs, as the next result shows.

\begin{defn}
Let $I$ be an interval of real numbers. Let $D\!:\ I^2\rightarrow\mathbb{R}$ be a moderate deviation function. Let~$\mathbb{W}$ be a $s\times t$ non-negative weighting matrix such that $w_{ij}\in[0,\infty)$ for every $i\in[s]$, $j\in[t]$, and $\mathbb{X}$ be a $s\times t$ matrix of real numbers such that $x_{ij}\in I$ for every $i\in[s]$, $j\in[t]$. The mapping $M_{D,\mathbb{W}}\!:\ I^{p\cdot q}\rightarrow I$ is said to be a \emph{weighted deviation-based matrix aggregation function} if and only if
\begin{align*}
M_{D,\mathbb{W}}(\mathbb{X})=
\frac{1}{2}
\big(\sup&\big\{y\in I\mid\sum_{i=1}^{s}\sum_{j=1}^{t}w_{ij}D(x_{ij},y)<0\big\}+ \nonumber\\
+\inf&\big\{y\in I\mid\sum_{i=1}^{s}\sum_{j=1}^{t}w_{ij}D(x_{ij},y)>0\big\}\big).
\tag{MDWX}\label{MDWX}
\end{align*}
The image $M_{D,\mathbb{W}}(\mathbb{X})$ of $\mathbb{X}$ is said to be a \emph{weighted deviation-based matrix aggregation}.
\end{defn}

\begin{prop}
Let $k\in[n]$ be an arbitrary but fixed integer. Let $\mathbb{B}_k^{\alpha\beta}$ be a $r\times r$ matrix defined by~\eqref{B_alphabeta}. Let $\mathbf{w}=(w_1,w_2,\dots,w_n)\in[0,\infty)^n$ be a non-negative weighting vector. Finally, referring to~\eqref{I_k} and~\eqref{MDX}, let $M_{D}\!:\ (I_k^{\alpha\beta})^{r\cdot r}\rightarrow\mathbb{R}$ be a deviation-based matrix aggregation function. Then, for every $k\in[n]$, a function $M_{D,\mathbf{w}}\!:\ (I_k^{\alpha\beta})^{r\cdot r}\rightarrow\mathbb{R}$ defined by
\begin{equation}
M_{D,w_k}(\mathbb{B}_{k}^{\alpha\beta})=
M_{D}(w_k\mathbb{B}_{k}^{\alpha\beta})=
M_{D}
\left(
\begin{array}{ccc}
w_k\cdot b_{11k}^{\alpha\beta}&\cdots &w_k\cdot b_{1rk}^{\alpha\beta}\\[6pt]
w_k\cdot b_{21k}^{\alpha\beta}&\cdots &w_k\cdot b_{2rk}^{\alpha\beta}\\[6pt]
\vdots&\ddots&\vdots\\[6pt]
w_k\cdot b_{r1k}^{\alpha\beta}&\cdots &w_k\cdot b_{rrk}^{\alpha\beta}\\
\end{array}
\right)
\label{MDw}
\end{equation}
is a weighted deviation-based matrix aggregation function.
\end{prop}

Let us finish the section with the definition of a $(p/r)\times (q/r)$ matrix $\mathbb{C}$ as an aggregational substitute of the $p\times q$ matrix~$\mathbb{A}$. Its elements shall be $n$-tuples of weighted deviation-based matrix aggregations.

\begin{defn}Let $I$ be an interval of real numbers. Let $(\mathbb{X}_1,\mathbb{X}_2,\dots,\mathbb{X}_n)$ be a vector of $s\times t$ matrices such that $x_{ij}\in I$ for every $i\in[s]$, $j\in[t]$. Let~$(\mathbb{W}_1,\mathbb{W}_2,\dots,\mathbb{W}_n)$ be a vector of $s\times t$ non-negative weighting matrices such that $w_{ij}\in[0,\infty)$ for every $i\in[s]$, $j\in[t]$. Let due to~\eqref{I_k}, for every $k\in[n]$, $D_k\!:\ I_k^2\rightarrow\mathbb{R}$ be a moderate deviation function. Then a vector $\mathbf{y}$ defined due to~\eqref{MDWX} as
\begin{equation}
\mathbf{y}=\big(
M_{D_1,\mathbb{W}_1}(\mathbb{X}_{1}),\,
M_{D_2,\mathbb{W}_2}(\mathbb{X}_{2}),\,
\dots,\,
M_{D_n,\mathbb{W}_n}(\mathbb{X}_{n})
\big)
\label{vector}
\end{equation}
is said to be a \emph{weighted deviation-based matrix aggregation vector}.
\end{defn}

\begin{prop}
Let $(\mathbb{B}_1^{\alpha\beta},\mathbb{B}_2^{\alpha\beta}, \dots,\mathbb{B}_n^{\alpha\beta})$ be a vector of $r\times r$ matrices defined by~\eqref{B_alphabeta}. Let $\mathbf{w}=(w_1,w_2,\dots,w_n)\in[0,\infty)^n$ be a non-negative weighting vector. Let due to~\eqref{I_k}, for every $k\in[n]$, $D\!:\ I_k^2\rightarrow\mathbb{R}$ be a (fixed) moderate deviation function. Then a vector $\mathbf{y}^{\alpha\beta}$ defined, due to~\eqref{MDw} and~\eqref{vector}, by
\begin{equation}
\mathbf{y}_\mathbf{w}^{\alpha\beta}=\big(
M_{D,w_1}(\mathbb{B}_1^{\alpha\beta}),\,
M_{D,w_2}(\mathbb{B}_2^{\alpha\beta}),\,
\dots,\,
M_{D,w_n}(\mathbb{B}_n^{\alpha\beta})
\big)
\label{abw_vector}
\end{equation}
is a weighted deviation-based matrix aggregation vector.
\end{prop}

\noindent{\sl Proof.} Straightforward.

\begin{defn}
Let~$\mathbb{A}$ be a $p\times q$ matrix of real $n$-tuples such that both $p$ and $q$ are divisible by a positive integer $r$, $r\neq 1$. Let~$(\mathbb{W}_1,\mathbb{W}_2,\dots,\mathbb{W}_n)$ be a vector of $s\times t$ non-negative weighting matrices such that $w_{ij}\in[0,\infty)$ for every $i\in[s]$, $j\in[t]$. Let due to~\eqref{I_k}, for every $k\in[n]$, $D_k\!:\ I_k^2\rightarrow\mathbb{R}$ be a moderate deviation function. Let, for every $\alpha\in[p/r]$, $\beta\in[q/r]$, $\mathbb{B}^{\alpha\beta}$ be a $r\times r$ matrix defined by~\eqref{b_alphabeta} and $\mathbf{y}^{\alpha\beta}$ be a weighted deviation-based matrix aggregation vector defined by
\begin{equation}
\mathbf{y}^{\alpha\beta}=\big(
M_{D_1,\mathbb{W}_1}(\mathbb{B}_1^{\alpha\beta}),\,
M_{D_2,\mathbb{W}_2}(\mathbb{B}_2^{\alpha\beta}),\,
\dots,\,
M_{D_n,\mathbb{W}_n}(\mathbb{B}_3^{\alpha\beta})
\big).
\label{abW_vector}
\end{equation}
A $(p/r)\times (q/r)$ matrix $\mathbb{C}=(c_{\alpha\beta})$ is said to be a \emph{weighted deviation-based aggregational substitute of~$\mathbb{A}$ with respect to $D_1,D_2,\dots,D_n$} if and only if
\begin{equation}
c_{\alpha\beta}=\mathbf{y}^{\alpha\beta}
\label{AofA}
\end{equation}
for every $\alpha\in[p/r]$, $\beta\in[q/r]$.
\end{defn}

\begin{prop}
Let~$\mathbb{A}$ be a $p\times q$ matrix of real $n$-tuples such that both $p$ and $q$ are divisible by a positive integer $r$, $r\neq 1$. Let $\mathbf{w}=(w_1,w_2,\dots,w_n)\in[0,\infty)^n$ be a non-negative weighting vector. Let due to~\eqref{I_k}, for every $k\in[n]$, $D\!:\ I_k^2\rightarrow\mathbb{R}$ be a (fixed) moderate deviation function. Then a $(p/r)\times (q/r)$ matrix $\mathbb{C}=(c_{\alpha\beta})$ defined due to~\eqref{abw_vector} by
\begin{equation}
c_{\alpha\beta}=\mathbf{y}_\mathbf{w}^{\alpha\beta}
\label{AofA}
\end{equation}
for every $\alpha\in[p/r]$, $\beta\in[q/r]$, is a weighted deviation-based aggregational substitute of~$\mathbb{A}$ with respect to $D$.
\end{prop}

\noindent{\sl Proof.} Straightforward.



\begin{rem}
\begin{enumerate}

\item Note that, although for the sake of simplicity we have made our analysis considering that we divide the original matrix into equal submatrices, the results extend straightforwardly also to the case when different sizes of submatrices are considered.

\item Observe that usual aggregation functions can be considered as functions which act over a $1\times n$ matrices of $1$-dimensional vectors and our approach, in particular, generalizes this case, too.

\end{enumerate}

\end{rem}
\begin{exam}\label{exam1}

Let $I$ be a real closed interval. Let~$\mathbb{A}$ be a $100\times 40$ matrix of real triples $a_{ij}$ such that
      $$a_{ij}=(a_{ij1},\,a_{ij2},a_{ij3}),\ i\in[100],\ j\in[40],\ \text{and}\ (a_{ij1},\,a_{ij2},a_{ij3})\in I^3.$$
      Let $r=2$. Thus, according to~\eqref{b_alphabeta}, $\alpha\in[100/2]=[50]$ and $\beta\in[40/2]=[20]$.

For $\varepsilon\geq 1$, define $D_\varepsilon\!:\ I^2\rightarrow \mathbb{R}$ by $$D_\varepsilon(x,y)=(x+\varepsilon)(y-x).$$
So, $D_\varepsilon$ is a deviation function, and the corresponding aggregation function of two variables $M_{D_\varepsilon}:\ I^2\rightarrow \mathbb{R}$ is given by
\begin{equation}
M_{D_\varepsilon}(u,v)=
       \frac{u(u+\varepsilon)+v(v+\varepsilon)}{u+v+2\varepsilon}
\label{eq_exam1}
\end{equation}
       for all $u,\, v\in\ I$ such that $u+v+2\varepsilon\neq 0$.

Let $\mathbf{w}=(w_1,w_2,w_3) \in [0,\infty)$ be a weighting vector. For all $k\in[3]$, $\alpha\in[50]$, $\beta\in[20]$, define, in accordance with~\eqref{B_alphabeta}, a $2\times 2$ matrix $\mathbb{B}_k^{\alpha\beta}$. Then a function $M_{D_\varepsilon,\mathbf{w}}\!:\ I^4\rightarrow \mathbb{R}$ given by
\begin{align*}
M_{D_\varepsilon,w_k}(\mathbb{B}_{k}^{\alpha\beta})&=
M_{D_\varepsilon}(w_k\mathbb{B}_{k}^{\alpha\beta})=
M_{D_\varepsilon}
\left(
\begin{array}{cc}
w_k b_{11k}^{\alpha\beta}&w_k b_{12k}^{\alpha\beta}\\[6pt]
w_k b_{21k}^{\alpha\beta}&w_k b_{22k}^{\alpha\beta}\\[6pt]
\end{array}
\right)=\\
&=\frac{\sum\limits_{i=1}^{2}\sum\limits_{j=1}^{2}w_k b_{ijk}^{\alpha\beta}(b_{ijk}^{\alpha\beta} +\varepsilon)}
{4w_k\varepsilon +\sum\limits_{i=1}^{2}\sum\limits_{j=1}^{2}w_k b_{ijk}^{\alpha\beta}}\addtocounter{equation}{1}\tag{\theequation}\label{eq:deviation_image}
\end{align*}
is the weighted deviation-based matrix aggregation function. And, finally due to~\eqref{abw_vector}, the resulting $50\times 20$ matrix $\mathbb{C}$ of\,\ $50\cdot 20=1\,000$ triples
$$\mathbf{y}_\mathbf{w}^{\alpha\beta}=
\big(M_{D_\varepsilon,w_1}(\mathbb{B}_{1}^{\alpha\beta}),
M_{D_\varepsilon,w_2}(\mathbb{B}_{2}^{\alpha\beta}),
M_{D_\varepsilon,w_3}(\mathbb{B}_{3}^{\alpha\beta})\big),\ \alpha\in[50],\,\beta\in[20],$$  is the weighted deviation-based aggregational substitute of~$\mathbb{A}$ with respect to $D_\varepsilon$.

Let as consider that $\mathbb{B}_{k}^{\alpha\beta}$ is an $r\times r$ matrix of real numbers. Then the corresponding weighted deviation-based matrix aggregation function $M_{D_\varepsilon,\mathbf{w}}\!:\ I^{r\cdot r}\rightarrow \mathbb{R}$ is given by
\begin{equation}
M_{D_\varepsilon,w_k}(\mathbb{B}_{k}^{\alpha\beta})=
\frac{\sum\limits_{i=1}^{r}\sum\limits_{j=1}^{r}w_k b_{ijk}^{\alpha\beta}(b_{ijk}^{\alpha\beta} +\varepsilon)}
{r^2w_k\varepsilon +\sum\limits_{i=1}^{r}\sum\limits_{j=1}^{r}w_k b_{ijk}^{\alpha\beta}}.\label{eq:deviation_image_for_r}
\end{equation}

Suppose both an $r\times r$ matrix $\mathbb{B}_{k}^{\alpha\beta}=\mathbb{B}$ (for the sake of simplicity ) of real numbers and a weight $w_k=w$ are fixed. Let us investigate the limit case for a parameter $\varepsilon$, i.e. $\varepsilon\to\infty$.

\begin{align*}
&\lim\limits_{\varepsilon\to\infty}
M_{D_\varepsilon,w_k}(\mathbb{B}_{k}^{\alpha\beta})=
\lim\limits_{\varepsilon\to\infty}M_{D_\varepsilon,w}(\mathbb{B})=\\
&=\lim\limits_{\varepsilon\to\infty}\frac{\sum\limits_{i=1}^{r}\sum\limits_{j=1}^{r}w b_{ij}(b_{ij} +\varepsilon)}
{r^2w\varepsilon +\sum\limits_{i=1}^{r}\sum\limits_{j=1}^{r}w b_{ij}}=
\lim\limits_{\varepsilon\to\infty}
\frac{w\sum\limits_{i=1}^{r}\sum\limits_{j=1}^{r}b_{ij}(b_{ij} +\varepsilon)}
{w\big(r^2\varepsilon +\sum\limits_{i=1}^{r}\sum\limits_{j=1}^{r} b_{ij}\big)}=\\
&=\lim\limits_{\varepsilon\to\infty}
\frac{\sum\limits_{i=1}^{r}\sum\limits_{j=1}^{r}b_{ij}^2+
	\varepsilon\sum\limits_{i=1}^{r}\sum\limits_{j=1}^{r}b_{ij}}
{r^2\varepsilon +\sum\limits_{i=1}^{r}\sum\limits_{j=1}^{r} b_{ij}}\
=\lim\limits_{\varepsilon\to\infty}
\frac{\sum\limits_{i=1}^{r}\sum\limits_{j=1}^{r}\frac{b_{ij}^2}{\varepsilon}+
	\sum\limits_{i=1}^{r}\sum\limits_{j=1}^{r}b_{ij}}
{r^2+\sum\limits_{i=1}^{r}\sum\limits_{j=1}^{r}\frac{b_{ij}}{\varepsilon}}=\\
&=\frac{\sum\limits_{i=1}^{r}\sum\limits_{j=1}^{r}b_{ij}}
{r^2}.
\end{align*}

Thus, regarding formula~\eqref{eq:deviation_image_for_r}, the limit assumption $\varepsilon\to\infty$ yields a standard arithmetic mean.
\end{exam}

\section{Some illustrative examples}\label{Exam_section}

\subsection{Illustrative example 1: Image processing}
Let us consider a color image of $p \times q$ pixels and suppose that the intensity of each pixel is given by three values: the first representing the intensity in the red channel; the second one representing the intensity in the blue channel; and the third one, representing the intensity in the green channel. In this case, the image can be identified with a matrix of size $p \times q$ whose entries are 3-tuples of real numbers.

\begin{figure}
	\centering
	\includegraphics[scale=0.75]{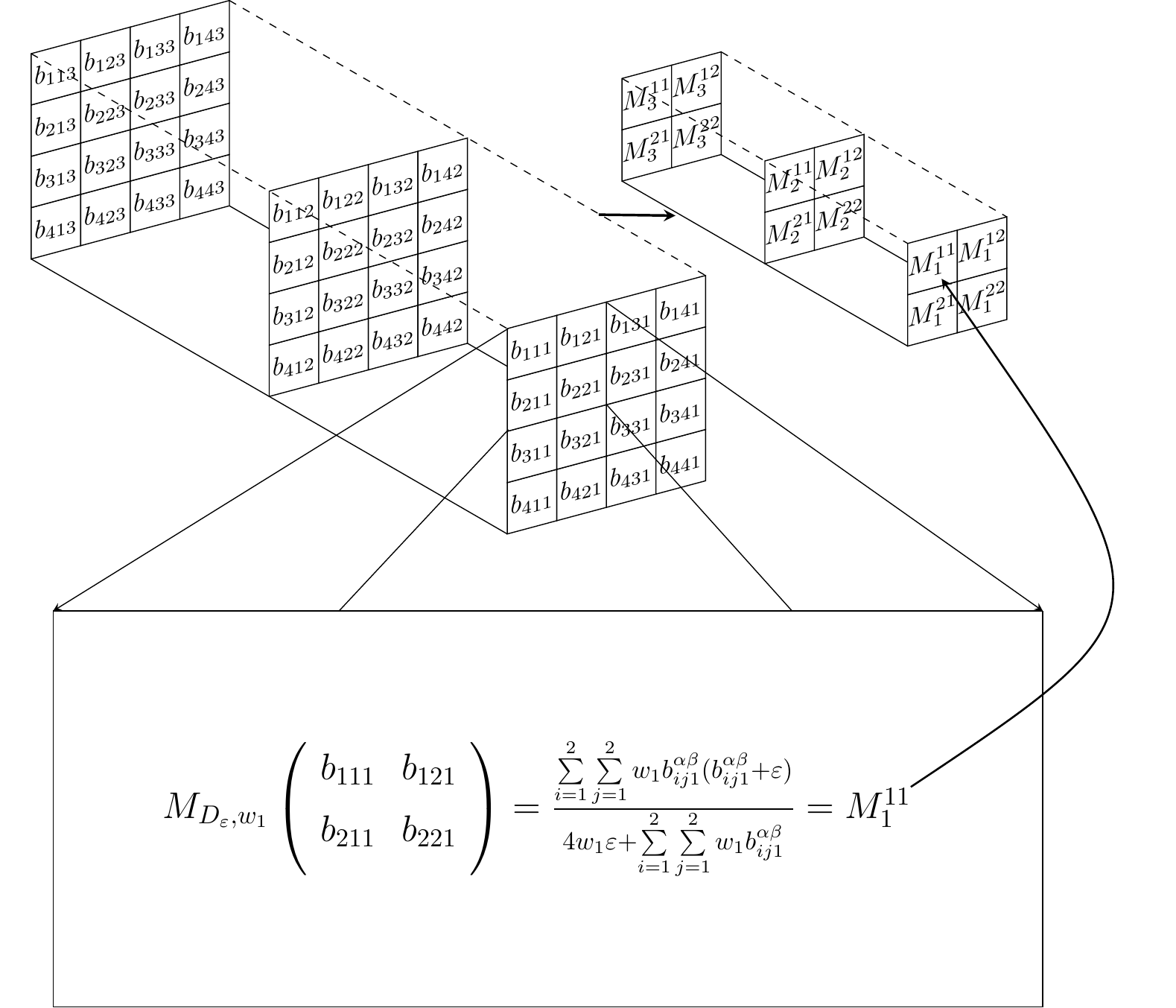}
	\caption{Example of the image reduction process to be performed. The values of each disjoint window $\mathbb{B}_k^{\alpha\beta}$ of the given image are aggregated into a single representative value $M_k^{\alpha\beta}$ using a weighted deviation-based matrix aggregation function.}
	\label{fig:image_reduction}
\end{figure}

Our objective will be to reduce a colour image with 3 channels of colour (RGB) of size $p\times q$ into a $(p/r)\times (q/r)$ one, using the function provided in Example~\ref{exam1}, Eq.~\eqref{eq_exam1}.  Figure~\ref{fig:image_reduction} shows the desired process. Ideally, the reduced version would allow us to recover the original image with as much fidelity as possible, by means of some magnification method.

For our experiments we will be using the test subset of the Berkeley Segmentation Dataset~\cite{PabloMichael}, which is comprised of 200 images of different sizes.

The process we will follow will consist on choosing a window size $r$ and applying Eq.~\eqref{eq:deviation_image_for_r} to reduce each of the disjoint $r \times r$ patches of our image to a single representative value. If $p$ or $q$ are not divisible by $r$, padding will be added to the necessary dimension \cite{PaddingPatch, PaddingImageProcessing}. 
We will use the vector of weights $\mathbf{w} = \mathbf{1} = (1, 1, 1)$, since we assume the same importance for each colour channel.

Finally, we will amplify the resulting image and compare it with the original one, using the Structural Similarity index (SSIM)~\cite{WangBovik}. This index has turned into the state of the art metric for image comparison, and it focuses on three different factors: the structure of the objects present in the images, their local luminance and their local contrast.
	
In order to compute the SSIM of two $N \times N$ windows $\mathbf{x}$ and $\mathbf{y}$ of two different images, the next equation can be used:
	
\begin{equation*}
SSIM(\mathbf{x},\mathbf{y})=\frac{(2\mu_x\mu_y+C_1)(2\sigma_{xy}+C_2)}{(\mu_x^2+\mu_y^2+C_{1})(\sigma_x^2+\sigma_y^2+C_2)}
\end{equation*}

where $\mu_x$ represents the mean intensity of the windows $\mathbf{x}$, $\sigma_x$ is its signal contrast, computed as the standard deviation of the intensities, $C_1$ and $C_2$ are constants included to avoid instabilities, and $\sigma_{xy}$ is computed as:

\begin{equation*}
\sigma_{xy}=\frac{1}{N-1}\sum_{i=1}^{N}(x_i-\mu_x)(y_i-\mu_y)
\end{equation*}

Note that the given equation only accounts for a single colour channel. In the case of multichannel images, SSIM indices are computed for each separate channel and averaged. Similarly, in order to calculate the SSIM index of two whole images, we will calculate the average of the SSIM index of each couple of disjoint windows.
	
Since our interest resides in the reduction phase of the algorithm, the magnification of the image will be performed by means of nearest neighbour interpolation (each pixel intensity will be replicated in a window of size $r \times r$), the simplest method which will allow us to evaluate the result in the most fair way.

\begin{figure}
	\centering
	\includegraphics[scale=0.75]{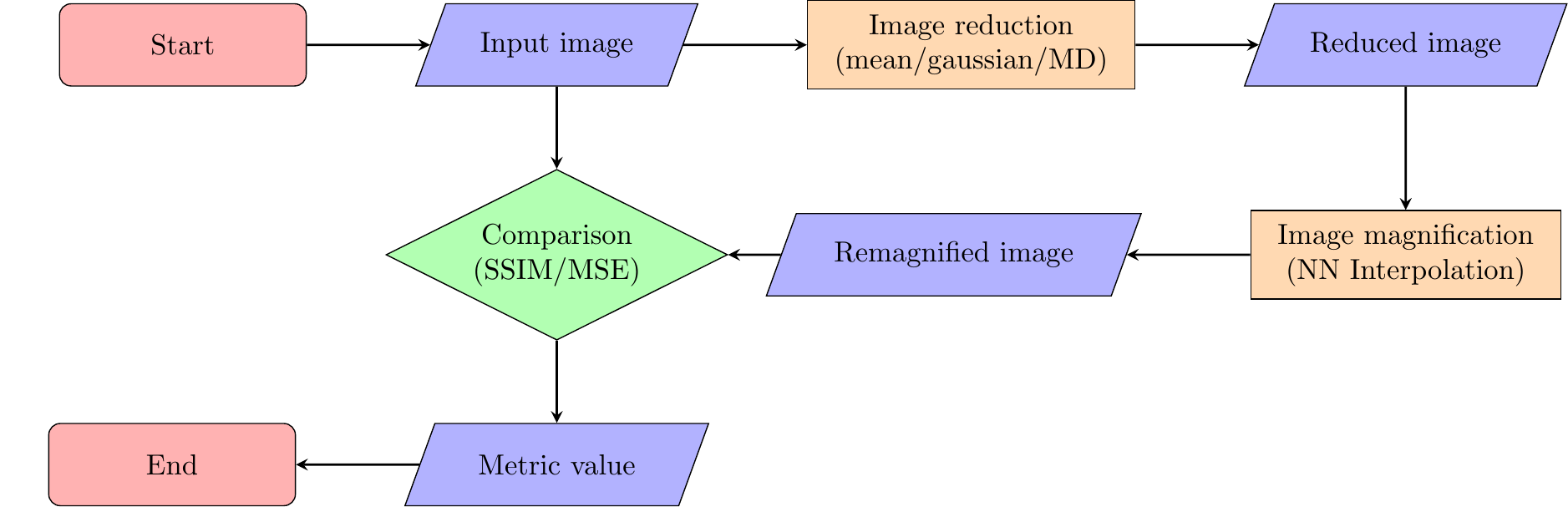}
	\caption{Process to be followed in our experimentation. A given image will be subsequently reduced and remagnified, and we will compare the resulting remagnified image with the initial one, by mean of some quality metric.}
	\label{fig:application1}
\end{figure}

We will compare the results obtained by our method, both with simple reduction methods such as mean, gaussian, median or geometric mean filter convolution as well as with other previously studied image reductions techniques. To this effect, we've tested both the effect of $K_\alpha$ operators such as the ones used in \cite{BustinceKAlpha} and centered OWA functions as in \cite{PaternainCOWA}. Finally, we compare our method with penalty functions as presented in~\cite{BeliakovPenalty}. In particular, for this last case we've used as possible aggregations the geometric mean, the minimum, the maximum, the average and the median. We've also set the Euclidean distance as penalty function. Figure~\ref{fig:application1} visually sums up the process to be performed.

\begin{figure}
	\centering
	\includegraphics[width=\textwidth]{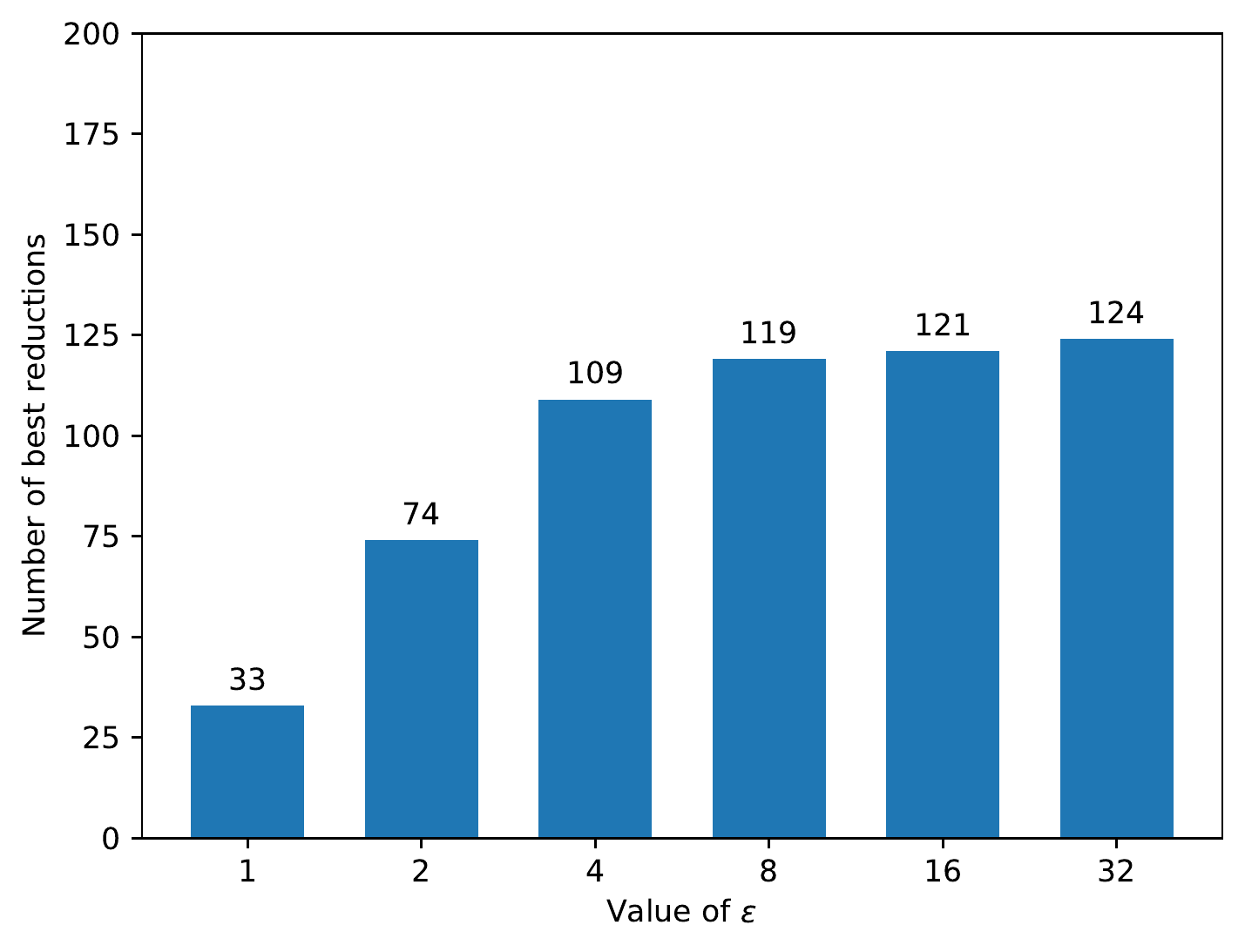}
	\caption{Number of best reductions based in the value of $\varepsilon$ (logarithmic scale for x axis)}
	\label{fig:epsilonevolutionabsolute}
\end{figure}

Figure~\ref{fig:epsilonevolutionabsolute} shows an interesting correlation between the value of $\varepsilon$ and the performance of the method, when using a value of $r=2$. As $\varepsilon$ is increased, the number of images for which the presented method obtains the best reduction increases. We suspect that this can be related with the fact that formula~\eqref{eq:deviation_image_for_r} approaches the behaviour of the arithmetical mean when $\varepsilon \to \infty$.

Based on the previous discovery, Table~\ref{tab:mean_ssim} presents the average SSIM and MSE (Mean Squared Error) obtained when taking into account windows of size $r = 2$ and big values of $\varepsilon$ (in particular $\varepsilon = 32$). Although the performance of all methods is pretty similar, the presented method works in pair with mean and gaussian reductions and slightly outperforms methods such as penalty functions with a much lighter computational load.

To this respect, it is important noting that penalty functions, when applied to $n$-dimensional data (3 colour channels in our example) require computing all the possible combinations with repetition of the selected functions, taken $n$ by $n$, in order to choose the best one with respect to a given distance, e. g. the euclidean distance. Unfortunately, that limits their applicability to problems with few dimensions or to the use of a small amount of aggregation functions at a time.

\begin{figure}
	\centering
	\includegraphics[width=\textwidth]{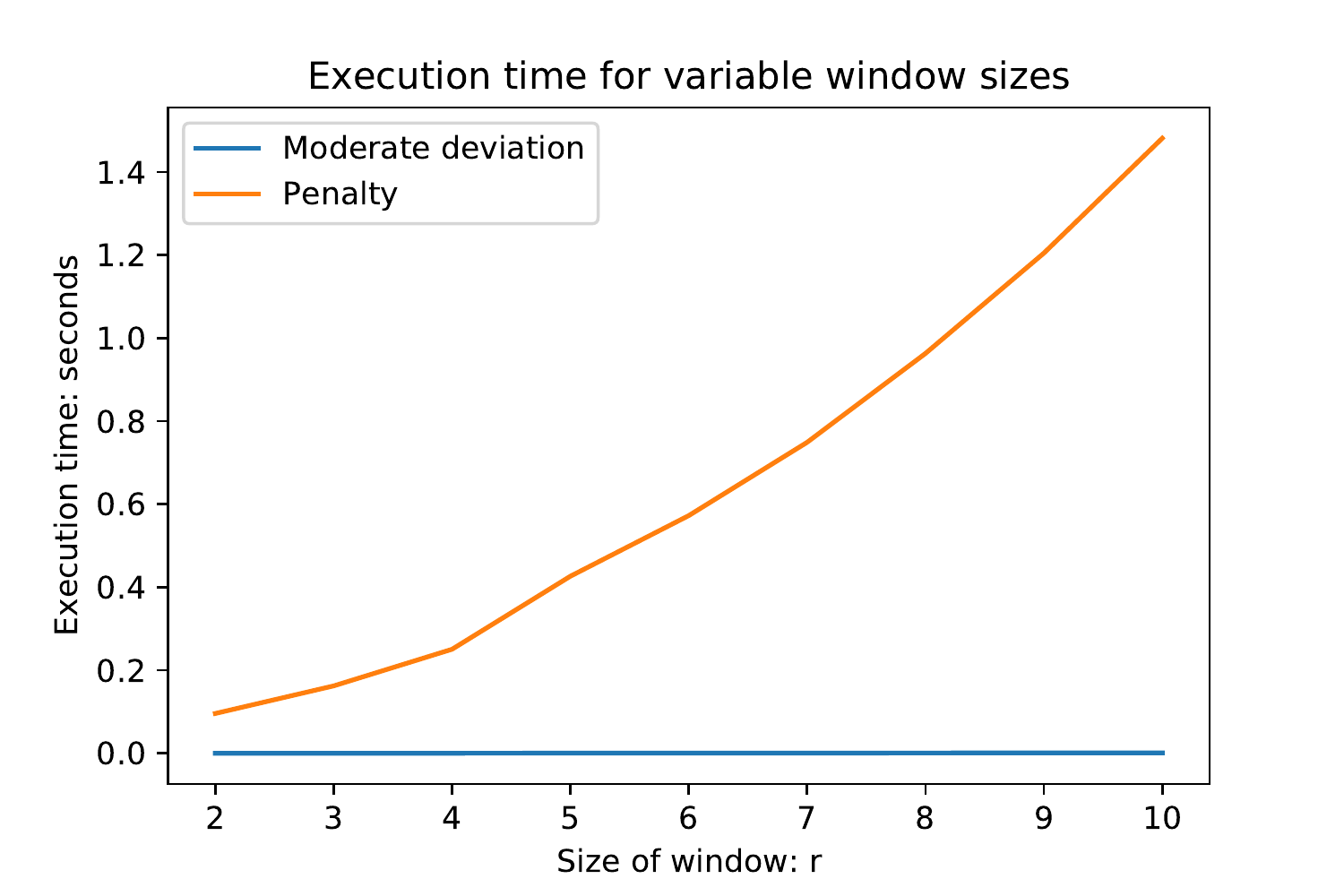}
	\caption{Computation time necessary for processing 500 windows of size $r \times r$ as $r$ increases, depending on the used method. The temporal cost of a penalty function is one of the limitations that moderate deviation functions can alleviate.}
	\label{fig:execution_time_complexity}
\end{figure}

Our method, on the other hand, computes a single value for each window. In order to test the temporal advantage that this presents, we have studied the temporal cost of reducing windows of size $r \times r$ of 3-tuples, as $r$ increases. Figure~\ref{fig:execution_time_complexity} gives a clear advantage to our method, whose temporal cost appears nearly constant when compared to that of the penalty function. We have also timed the processing time needed to reduce all the images on the Berkeley dataset using both methods. While the penalty function requires 190.75 seconds of processing time, our method finishes the job in just 4.03 seconds, yielding a 47 times faster execution.

Since the best values are obtained for this window size, which is the default one used for the pooling phase of Convolutional Neural Networks, we will also test the performance of this method in the field of Deep Learning.

\begin{table}
	\centering
\caption{Mean metric value obtained for the Berkeley Dataset with each of the methods.}
	\begin{tabular}{| c | c | c |}
		\hline
		Method & SSIM & MSE \\
		\hline
		MD & $\mathbf{0.900740}$ & $\mathbf{0.002291}$ \\
		Mean & 0.900734 & $\mathbf{0.002291}$ \\
		Median & 0.898354 & 0.002399 \\
		Gaussian & 0.900734 & $\mathbf{0.002291}$\\
		Geometric mean & 0.897037 & 0.002456\\
		$k_{0.25}$ & 0.885056 & 0.0032\\
		$k_{0.5}$ & 0.896107 & 0.002383\\
		$k_{0.75}$ & 0.885444 & 0.003299\\
		cOWA & 0.898464 & 0.002338\\
		Penalty & 0.900437 & 0.002295\\
		\hline
	\end{tabular}
	\label{tab:mean_ssim}
\end{table}


\subsection{Illustrative example 2: deep learning}

In Deep Learning, we call convolutional neural network to an artificial neural network architecture composed of a succession of convolution and pooling layers~\cite{LecunCNN}.

In the field of image processing, the convolution of several ``filters" over a given image is a common technique used for solving problems such as thresholding, image reduction or edge detection. However, setting the values of these filters a priori can be a difficult task that requires lots of experience and parameter tweaking. Convolutional Neural Networks (CNNs) automate this job, by means of defining an arbitrary number of convolution filters (or kernels) of random values, which are modified via an iterative optimization process. These filters are used to extract significative features of a given image, that are used for its classification.

\begin{figure}
	\centering
	\includegraphics[scale=0.75]{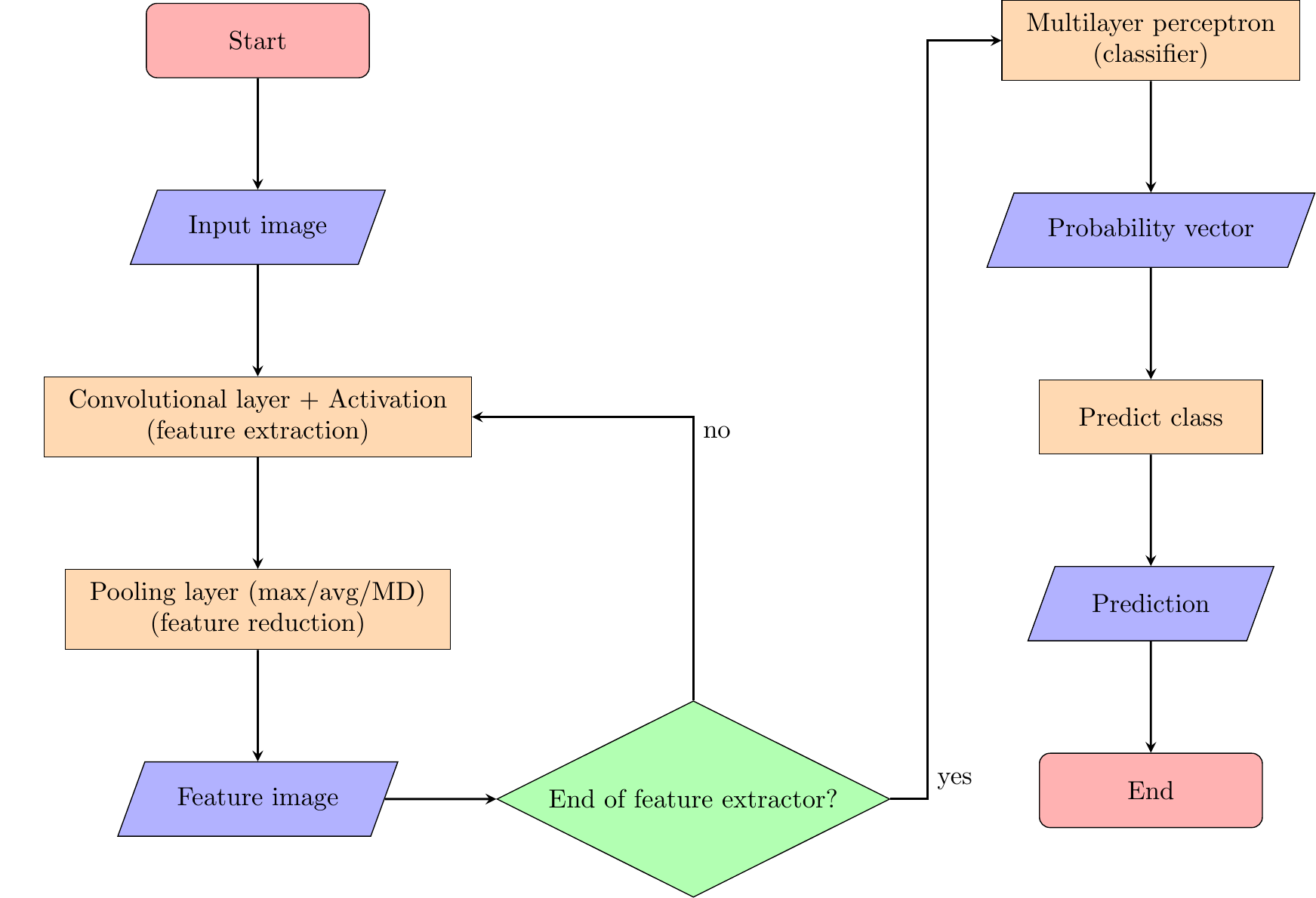}
	\caption{Process of a classical Convolutional Neural Network. The model receives an input image and predicts the class to which it belongs. The first part of the model extracts and compresses visual features that are afterwards used as input for a classifier.}
	\label{fig:application2}
\end{figure}

This behaviour implies that CNNs are composed of two interconnected parts: The first one takes care of the feature extraction process, receiving an image as input and returning a representative feature vector as output; the second acts as a classifier that receives this feature vector and determines the class to which the initial image belongs. The complete process is exemplified by Figure~\ref{fig:application2}.

The first part is, at the same time, composed of a series of ``layers" connected in a sequential way, so that the output of a layer acts as the input for the following. Although in the last few years lots of new layers have been introduced~\cite{Ioffe} \cite{Srivastava}, the two main ones, common to every CNN architecture, are convolutional and pooling layers.

A convolutional layer consists of a set of $N$ kernels of dimension $S \times S \times D$. Their values (also known as weights) are randomly initialized, and modified iteratively using some optimization algorithm, usually some variant of the gradient descent optimization~\cite{Ruder}.

When a convolutional layer receives a three dimensional image as input, each of these kernels are shifted over the input image, performing a matrix multiplication between the values covered by the kernel in each moment and the weights of the kernel itself. A non linear activation function is applied to the result of the previous multiplication, in order to being able to obtain non linear outputs for the network, before shifting the kernel to the next position and going on with the process. The shift of each of these kernels over the whole image produces a different feature image, that is stacked with the rest before sending the resulting three dimensional feature image to the next layer.

Pooling layers are placed after convolutional layers, in order to reduce the dimensionality of feature images progressively. To achieve this, pooling layers can be thought of as dividing the input image into disjoint matrices of size $n \times  n$ (usually $2 \times 2$ or $3 \times 3$) and aggregating the values of each one, most commonly using the maximum or the average. In the last years, however, new pooling functions have been proposed: in~\cite{stochasticPool} the authors present a pooling method which relies on a probabilistic procedure to select one of the values of the pooling window; in~\cite{forcen2020pooling} the OWA function is presented as a candidate for the pooling operation.

In our experimentation we have used the MD pooling instead, a new pooling layer that makes use of our expressions, in particular Eq.~\eqref{eq:deviation_image_for_r}, in a way such as the one described by Figure~\ref{fig:image_reduction}. Since these expressions take into account a weight vector $W$, our method allows for the introduction of a learnable parameter that weights each of the feature image channels differently. We refer to this variant of MD pooling as LMD pooling. However, we always have the possibility of setting $W=\mathbf{1}$, in the same way that we did in the image reduction example.

\begin{figure}
	\centering
	\includegraphics[width=\textwidth]{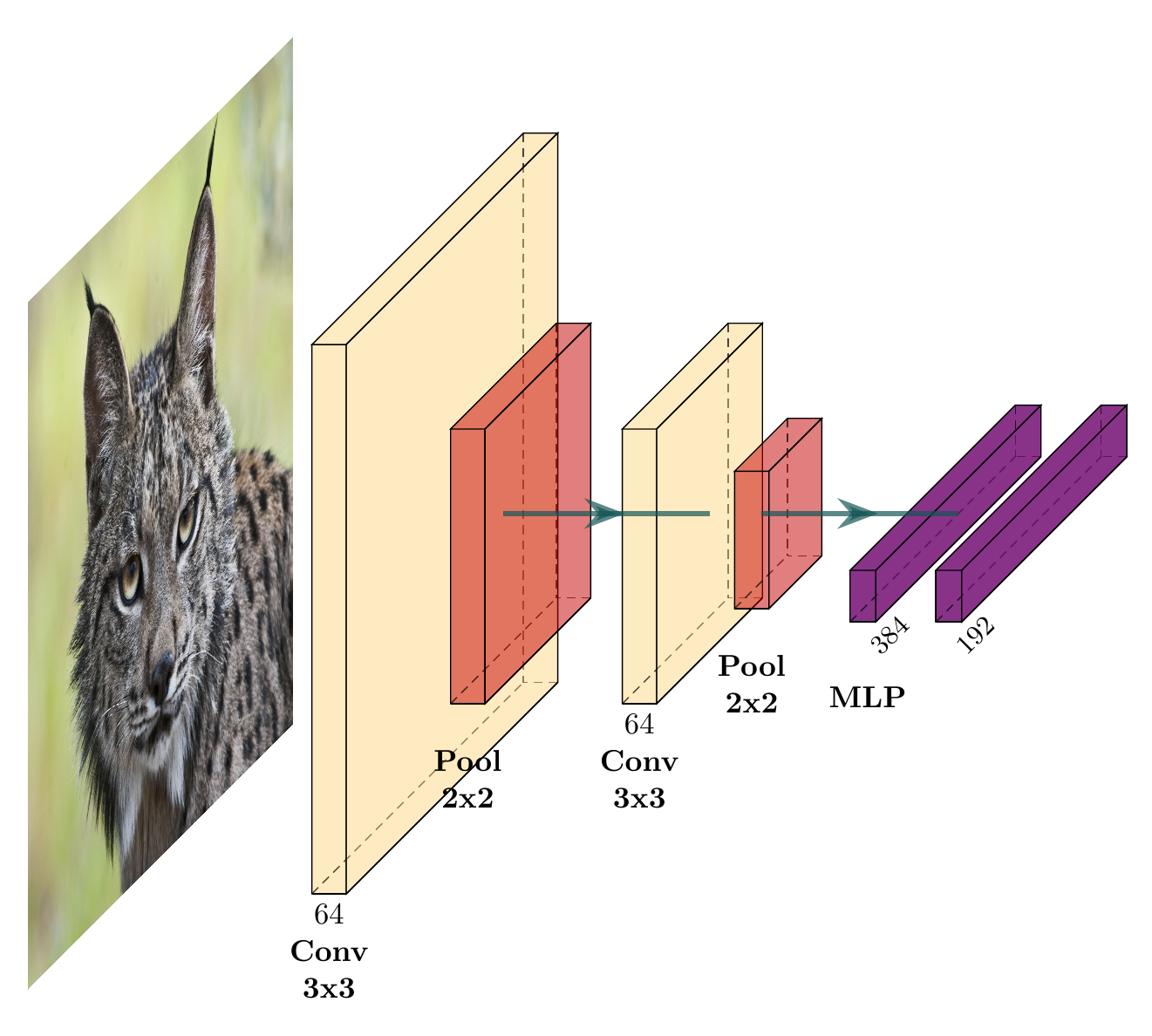}
	\caption{Network architecture used in the experimentation.}
	\label{fig:network}
\end{figure}

We have compared the performance of a simple test architecture (Figure~\ref{fig:network} and Table \ref{tab:net}) when using the MD pooling layers, against the usual maximum and average ones as well as the stochastic pooling and the OWA pooling.

In Table \ref{tab:layer_combinations} we show our results for the Cifar 10 dataset~\cite{Krizhevsky} after training for 256 epochs with different pooling layers combinations. The CIFAR10 dataset is composed of 60000 32x32 images, evenly distributed among 10 different classes. 50000 of those images form the train set, while the remaining 10000 are the ones reserved for testing purposes.

As it can be seen, the models which just use one type of layer perform similarly to max, OWA and stochastic poolings, although being fairly outperformed by the average pooling. However, when we combine a LMD layer with one of the other aggregations, the results show a clear improvement, obtaining the best accuracy for the model which uses LMD as first pooling function and the average for the second. This implies that the combination of different types of pooling layers may present an interesting strategy for training deeper models. This study, however, falls out of the scope of the current paper.

Similarly to the example on image reduction, we have also studied the influence of the parameter $\varepsilon$ in the accuracy of the model. Table \ref{tab:eps_impact} shows the results obtained after training on Cifar 10, both with the fixed and learnt W. Although some values of $\varepsilon$ greater than 1 offer better accuracy results in the fixed case, we don't find such a direct relationship as in the previous application. We have to take into account that during training, thousands of parameters are tuned at the same time, and their evolution will vary based in the rest of parameters of the network (namely $\varepsilon$). One could try to set $\varepsilon$ as an additional trainable parameter of the network, but the $\varepsilon \ge 1$ constraint poses a problem for its learning.

\begin{table}
	\centering
    \caption{Accuracy for each of the different pooling combinations tested.}
	\begin{tabular}{| cc | c |}
		\hline
		Pool 1 & Pool 2 & Accuracy \\
		\hline
		Max & Max & 79.6\%\\
		Avg & Avg &  82.8\%\\
		MD & MD & 79.6\%\\
		LMD & LMD & 78.8\%\\
		LMD & Max & 81.6\%\\
		Max & LMD & 81.6\%\\
		LMD & Avg & $\mathbf{83.0\%}$\\
		Avg & LMD & 80.1\%\\
		OWAQ & OWAQ & 79.9\%\\
		Stochastic & Stochastic & 80.3\%\\
		\hline
	\end{tabular}
	\label{tab:layer_combinations}
\end{table}

\begin{table}
	\centering
    \caption{Accuracy obtained for CIFAR10 dataset for different values of $\varepsilon$.}
	\begin{tabular}{| c | c | c |}
		\hline
		$\varepsilon$ & $\mathbf{W} = \mathbf{1}$ & $\mathbf{W}$ learnt \\
		\hline
		1 & 79.8\% & 80.1\% \\
		2 & $\mathbf{82.2\%}$ & 81.5\% \\
		4 & 79.0\% & 79.0\% \\
		8 & 78.7\% & 79.8\% \\
		16 & 78.1\% & 79.6\% \\
		32 & 80.5\% & 78.3\% \\
		\hline
	\end{tabular}
	\label{tab:eps_impact}
\end{table}

\begin{table}
	\centering
    \caption{Network architecture detailed used in the experimentation.}
	\begin{tabular}{ccc}
		\hline
		Layer Name & Output Size & Kernel Size \\
		\hline
		Conv1 & 32x32x64 & 3x3 \\
		ReLu1 & 32x32x64 & - \\
		Pool1 & 16x16x64 & 2x2 \\
		
		Conv2 & 16x16x64 & 3x3 \\
		ReLu2 & 16x16x64 & - \\
		Pool2 & 8x8x64 & 2x2 \\
		Flatten & 384x192 & - \\
		MLP1 & 192xC & - \\
		MLP2 &C & - \\
		\hline
	\end{tabular}
	\label{tab:net}
\end{table}

\subsection{Illustrative example 3: Decision making}

This example is devoted to the algorithm application in multi-expert decision making problem.

Let us consider a collection of $n$ experts $E=\{e_1,\dots,e_n\}$. Let $F=\{F_1,\dots,F_p\}$ represent a set of alternatives such that $|F|>1$. Suppose that each expert $e_k$, $k\in[n]$, provides his/her preference resulting in a $p\times p$ matrix $\mathbb{X}_k$ of real numbers $x_{ijk}\in[0,1]$, $i\in[p]$, $j\in[p]$, such that $x_{ijk}$ represents the expert's $e_k$ preference of alternative $F_i$ over alternative $F_j$. Let us also assume that for every $k \in [n]$ and for every $i \in [p]$, $x_{iik}$ has a fixed predetermined value.

The usual procedure to solve this kind of problems is to fuse all the preference matrices into one single collective matrix and then apply some algorithm to get the final decision.  But, in order to apply our developments, observe that we have a vector $(\mathbb{X}_1,\mathbb{X}_2,\dots,\mathbb{X}_n)$ of matrices. This vector can be understood as a three dimensional matrix $\mathbb{X}$ of dimensions $p\times p\times n$, which combines in one single multidimensional matrix the preference matrices provided by each of the experts. Let us call this matrix multi-dimensional preference matrix (MD preference matrix).

Let us now fix two indexes of the MD preference matrix $\mathbb{X}$, i.e. let $i\in[p]$, $j\in[p]$ be regarded as constants. If we consider the vector composed of the preference of each expert's preference of alternative $i$ over alternative $j$, this yields a vector $$\mathbf{x}_{ij\B}=(x_{ij1},x_{ij2},\dots,x_{ijn})$$ of preferences of the alternative $F_i$ over the alternative $F_j$ throughout all experts. Let us call it a \emph{preference vector}.

Suppose that each expert $e_k\in E$ has a weight $w_k\in[0,\infty]$ which yields a weighting vector $\mathbf{w}=(w_1,w_2,\dots,w_k)$. Since each $\mathbf{x}_{ij\B}$ represents the experts' collective opinion on the same subject, in order to get the collective matrix it is enough to  fuse the $n$-tuple into one single representative, taking into account the weights assigned to each of the experts. Let us do it applying a weighted deviation-based aggregation function (see~\eqref{MDX} and~\cite[Definition~4.1]{Decky}).

Let $D\!:\ [0,1]^2\rightarrow[0,1]$ be a moderate deviation function. Define a function $M_{D,\mathbf{w}}\!:\ [0,1]^n\rightarrow[0,1]$ by
\begin{align*}
M_{D,\mathbf{w}}(\mathbf{x}_{ij\B})&= M_{D,\mathbf{w}}(x_{ij1},x_{ij2},\dots,x_{ijn})=\\
&=
\frac{1}{2}
\big(\sup\big\{y\in [0,1]\mid\sum_{k=1}^{n}w_k D(x_{ijk},y)<0\big\}\,+\\
&\quad\,\, +\inf\big\{y\in [0,1]\mid\sum_{k=1}^{n}w_k D(x_{ijk},y)>0\big\}\big).\\
\end{align*}
for every vector $\mathbf{x}_{ij\B}$. Then $M_{D,\mathbf{w}}$ is  a weighted deviation-based aggregation function.

Further, let $\mathbb{C}$ be a $p\times p$ matrix such that
$$c_{ij}=M_{D,\mathbf{w}}(\mathbf{x}_{ij\B})\quad\text{for all}\ i\in[p],\ j\in[p].$$
Then an element $c_{ij}$ of~$\mathbb{C}$ is said to be a \emph{weighted deviation-based aggregational preference of $F_i$ over $F_j$} (abbreviated as \emph{aggregational preference}), and the resulting matrix $\mathbb{C}$ is said to be a \emph{weighted deviation-based preference collective matrix for~\,$\mathbb{X}$} (abbreviated as \emph{collective matrix}).

Conclude the algorithm by obtaining an aggregational preference column (vector) which enables us to opt for one of $F$-alternatives. Referring to the above described matrix $\mathbb{C}$ as a collective matrix for~\,$\mathbb{X}$, let $\mathbf{d}=(d_1,d_2,\dots,d_p)$ be a vector of real numbers given by
\begin{align}
d_i=M_D(\mathbf{c}_{i\B})&= M_D(c_{i1},c_{i2},\dots,c_{ip})=\nonumber\\
&=
\frac{1}{2}
\big(\sup\big\{y\in [0,1]\mid\sum_{j=1}^{p} D(c_{ij},y)<0\big\}\,+\nonumber\\
&\quad\,\, +\inf\big\{y\in [0,1]\mid\sum_{j=1}^{p} D(c_{ij},y)>0\big\}\big).
\label{preference_column}
\end{align}
for every vector $\mathbf{c}_{i\B}$, $i\in[p]$. Then the vector $\mathbf{d}$ is said to be a \emph{deviation-based aggregational preference column of~\,$\mathbb{X}$}.

\begin{figure}
		\centering
		\includegraphics[scale=0.75]{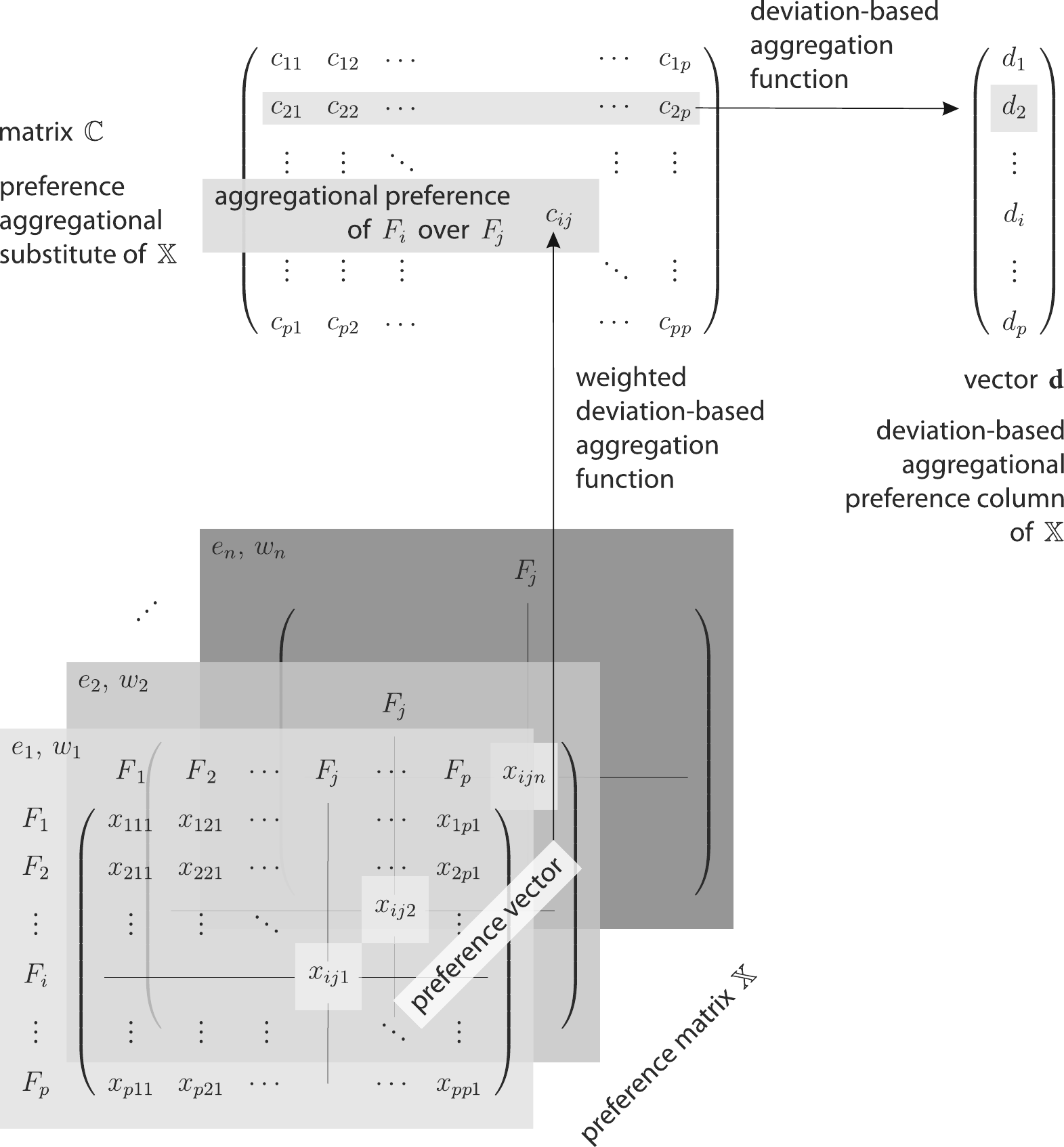}
		\caption{The algorithm application in multi-expert decision making problem yielding a \emph{deviation-based aggregational preference column}: step by step}
		\label{fig:algorithm}
\end{figure}

For the sake of better understanding the corresponding algorithm is described via Figure~\ref{fig:algorithm}.\medskip

Applying expression~\eqref{eq_exam1} in Example~\ref{exam1}, a corresponding formula to define a collective matrix for~$\mathbb{X}$ is given by
$$\displaystyle c_{ij}=M_{D,\mathbf{w}}(\mathbf{x}_{ij\B})=\frac{\sum\limits_{k=1}^{n}w_k\cdot x_{ijk}(x_{ijk}+\varepsilon)} {\varepsilon\sum\limits_{k=1}^{n}w_k+\sum\limits_{k=1}^{n}(w_k\cdot x_{ijk})}$$
for all $i\in[p]$, $j\in[p]$, and subsequently a deviation-based aggregational preference column of~\,$\mathbb{X}$ is given by
$$\displaystyle d_i=M_D(\mathbf{c}_{i\B})=\frac{\sum\limits_{j=1}^{p} c_{ij}(c_{ij}+\varepsilon)} {\varepsilon q+\sum\limits_{j=1}^{p}c_{ij}}$$
for all $i\in[p]$.

\section{Conclusion and future work}

In this paper we have provided a way to extend the use of moderate deviation functions in order to fuse multivalued data. More specifically, we have shown how to reduce a matrix of multivalued values using these functions, without the requirements imposed by other methods such as penalty functions. As this is a problem very common in different fields, we have illustrated the possible use of our proposal in problems in image processing, deep learning and decision making, fields in which, for different reasons, the use of other methods such as penalty functions is not fully suitable.

We have shown that the presented method constitutes an efficient alternative to these functions, which require extensive computational resources. Furthermore, they have the potential of being applicable to new fields via the study of different deviation functions suited for their specific use cases.

This work has provided a first approach to the notion of moderate deviation functions in a multivalued setting. The analysis of such functions would require much more space, and in particular, it could be of interest to consider the possibility of mixing different components of the multivalued inputs in order to get the final output, so that different pieces of information in input data are related to each other.

\section{Acknowledgments}

The research done by Humberto Bustince, Iosu Rodr\'\i guez Mart\'\i nez and Javier Fumanal Idocin has been funded by the project    (PID2019-108392GB-I00: 3031138640/AEI/10.13039/501100011033). The work of Martin Pap\v{c}o was supported by the Slovak Research and Development Agency under the contract No.~APVV-16-0073.


\bibliographystyle{elsarticle-num}
\bibliography{bibliography} 

\end{document}